\documentclass[a4paper,fleqn]{cas-dc}
\usepackage[authoryear]{natbib}
\usepackage{msc}
\usepackage{soul}
\usepackage{dutchcal}
\usepackage{framed,multirow}
\usepackage{algorithm,algcompatible,amsmath,import}
\usepackage{times,float,array,epsfig,xspace}
\usepackage{subfig}
\usepackage{graphicx}
\usepackage{amssymb}
\usepackage{verbatim}
\usepackage{longtable}
\usepackage{latexsym}

\usepackage{url}
\usepackage{xcolor}
\usepackage{setspace}
\usepackage[toc,page]{appendix}
\usepackage[flushleft]{threeparttable}
\newcommand{\etal}{\textit{et al.}}
\algnewcommand\INPUT{\item[\textbf{Input:}]}%
\algnewcommand\OUTPUT{\item[\textbf{Output:}]}
\definecolor{newcolor}{rgb}{.8,.349,.1}

\def\tsc#1{\csdef{#1}{\textsc{\lowercase{#1}}\xspace}}
\tsc{WGM}
\tsc{QE}
\tsc{EP}
\tsc{PMS}
\tsc{BEC}
\tsc{DE}

\begin{document}
\let\WriteBookmarks\relax
\def\floatpagepagefraction{1}
\def\textpagefraction{.001}
% Short title
\shorttitle{MC-ViViT: Multi-branch Classifier-ViViT to Detect Mild Cognitive Impairment in Older Adults Using Facial Videos}
% Short author
\shortauthors{J.Sun \etal}
% Main title of the paper

\title [mode = title]{MC-ViViT: Multi-branch Classifier-ViViT to Detect Mild Cognitive Impairment in Older Adults Using Facial Videos}

% author 1
\author[1]{Jian Sun}[type=editor,
                     %role=Researcher,
                     style=english,
                     orcid=0000-0002-9367-0892]
\fnmark[1]
% Email id of the first author
\ead{Jian.Sun86@du.edu}
% URL of the first author
\ead[url]{https://sites.google.com/view/sunjian/home}

% Address/affiliation
\affiliation[1]{organization={Department Of Computer Science, University of Denver},
            addressline={2155 E Wesley Ave}, 
            city={Denver},
            state={Colorado},
            postcode={80210}, 
            country={United States of America}}

\author[2]{Hiroko H. Dodge}[orcid=0000-0001-7290-8307]
\fnmark[2]
\affiliation[2]{organization={Department Of Neurology at Harvard Medical School, Harvard University},
            addressline={Massachusetts General Hospital, 55 Fruit St}, 
            city={Boston},
            state={Massachusetts},
            postcode={02114}, 
            country={United States of America}}
% Email id of the second author
\ead{hdodge@mgh.harvard.edu}
% URL of the second author
\ead[url]{https://dodgelab.wixsite.com/dodge-lab}

\author[3]{Mohammad H. Mahoor}[orcid=0000-0001-8923-4660]
\cormark[1]
\affiliation[3]{organization={Department Of Computer Engineering, University of Denver},
            addressline={2155 E Wesley Ave}, 
            city={Denver},
            state={Colorado},
            postcode={80210}, 
            country={United States of America}}
\fnmark[3]
\cortext[cor1]{Corresponding author: Dr.Mohammad H. Mahoor \\
Tel: +1-303-871-3745 | Fax: +1-303-871-2194}
% Email id of the second author
\ead{mohammad.mahoor@du.edu}
% URL of the second author
\ead[url]{http://mohammadmahoor.com}

% Here goes the abstract
\begin{abstract}
Deep machine learning models including Convolutional Neural Networks (CNN) have been successful in the detection of Mild Cognitive Impairment (MCI) using medical images, questionnaires, and videos. This paper proposes a novel Multi-branch Classifier-Video Vision Transformer (MC-ViViT) model to distinguish MCI from those with normal cognition by analyzing facial features. The data comes from the I-CONECT, a behavioral intervention trial aimed at improving cognitive function by providing frequent video chats. MC-ViViT extracts spatiotemporal features of videos in one branch and augments representations by the MC module. The I-CONECT dataset is challenging as the dataset is imbalanced containing Hard-Easy and Positive-Negative samples, which impedes the performance of MC-ViViT. We propose a loss function for Hard-Easy and Positive-Negative Samples (HP Loss) by combining Focal loss and AD-CORRE loss to address the imbalanced problem. Our experimental results on the I-CONECT dataset show the great potential of MC-ViViT in predicting MCI with a high accuracy of 90.63\% accuracy on some of the interview videos. 
\end{abstract}

\begin{keywords}
Deep Learning, \sep Facial Expression Features, \sep Inter- and Intra-class imbalance, \sep Mild Cognitive Impairment, \sep Multi-branch Classifier, \sep Transformer, ViViT.
\end{keywords}

\maketitle

\section{Introduction} \label{sec:1}

Alzheimer’s disease (AD) and related dementias (ADRD) are national public health issues with pervasive challenges for older adults, their families, and caregivers. ADRD is not only ranked as the sixth-leading cause of death in the U.S., but the COVID-19 pandemic has increased the number of deaths among those with AD~\citep{S1_5-AD_Facts_2021}. Symptoms of AD/ADRD usually begin with Mild Cognitive Impairment (MCI) and include early onset memory loss, cognitive decline, and impairments with verbal language and visual/spatial perception. Approximately 12-18\% of people age 60 or older are living with MCI in the U.S. Although older adults with MCI have the ability to independently perform most daily living activities such as eating, shopping, and bathing without help, the National Institute on Aging (NIA) estimated that 10 to 20\% of subjects with MCI will develop dementia over a one-year period~\citep{S1_6-MCI_Intro_2021}.

Magnetic Resonance Imaging (MRI) \& Positron Emission Tomography (PET) scanning and neuropsychic examination are often used for identifying those with AD and dementia. These methods prove to be challenging for early MCI individuals given that the brain’s structural changes and cognitive test results at this point might be harder to differentiate from those with normal cognitive aging and the invasive nature of these assessments~\citep{S1_7-MCI_Dialogue_2020}. Using accurate, non-invasive, and cost-efficient diagnostic technology has the potential to bolster the early detection of MCI and AD. A 2019 study proposed that early detection of AD would decrease healthcare costs, and improve quality of life~\citep{S1_9-BEAT-IT_2019}. Studies have shown that MCI can affect the patterns of speech, language, and face-to-face communication in older adults \citep{S1_8-MCI_CrossModel_2023}. There are several studies on using traditional Machine Learning (ML) approaches such as Decision Tree~\citep{S2_1_5-SVM_DTree_2020,S2_1_7-survey1_2020} and K-means clustering~\citep{S2_1_7-survey1_2020} to detect AD with a human using collected demographic information~\citep{S2_1_5-SVM_DTree_2020} and MRI~\citep{S2_1_7-survey1_2020}. Other researchers used Cross-model Augmentation for automated detection of MCI from normal cognition (NC) using speech and language patterns~\citep{S1_8-MCI_CrossModel_2023}. However, there is fewer work on using ML models and particularly Deep Learning models for facial video analysis and then automated detection of MCI using visual features. This paper presents our recent work on developing and using Transformer-based models for the detection of MCI using facial videos collected in the Internet-Based Conversational Engagement Clinical Trial (I-CONECT) Study project\footnotemark{}\footnotetext{Website: \url{https://www.i-conect.org/}.}~\citep{S1_10-I-CONECT-1_2019,S1_12-I-CONECT-3_2021,S1_11-I-CONECT-2_2022} (Clinicaltrials.gov \#: NCT02871921). I-CONECT Study is a randomized controlled behavioral intervention trial aimed to enhance cognitive functions by providing frequent social interactions using video chats. Semi-structured 30-minute conversations with interviewers were provided 4 times per week for 6 months to older adults. The study was funded by the National Institute on Aging (NIA).

\begin{figure*}[ht]
\begin{center}
\includegraphics[width=0.85\linewidth,height=0.26\textheight]{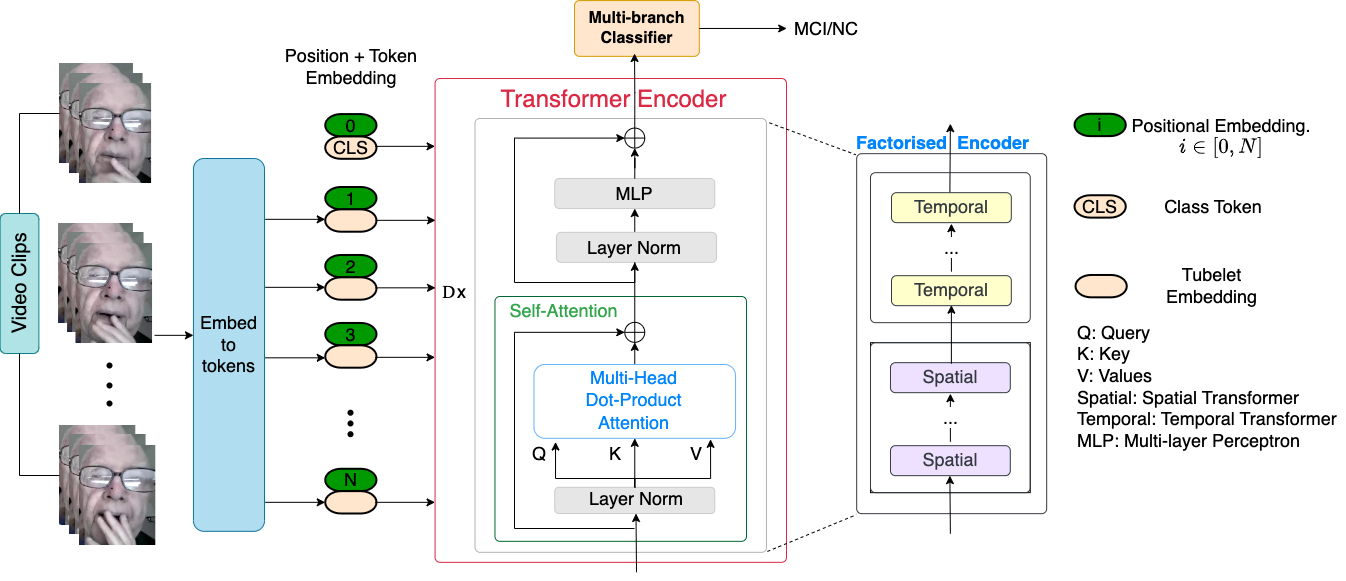}
\end{center}
\vspace{-0.3cm}
\caption{The structure of the proposed MC-ViViT. D is the model depth. It represents the layer number of MC-ViViT, which is the total number of Spatial Transformers and Temporal Transformers. Layer Norm is Layer Normalization.} MC-ViViT takes tubelet embedding to divide the video clips into cubes and loads these cubes into Transformer Encoder sequentially. Then, Transformer Encoder takes Factorised Encoder to extract spatio-temporal features. Finally, the output features go into Multi-branch Classifier to finish classification.
\label{fig:m-struct}
\vspace{-0.4cm}
\end{figure*}

Facial videos contain spatial features like facial expressions and head poses. It can also capture facial expression change, track eye gaze motion, head pose movement, and lips activity, which are the temporal features and the key patterns in verbal and non-verbal communication ~\citep{S2_1_22-Interact_2019,S2_1_19-EyeGaze_2020}. We hypothesize that facial videos collected during face-to-face communication in social settings can provide rich information for the detection of MCI from NC subjects. To capture facial features out of facial videos, we develop a variant of the Video Vision Transformer-based model (ViViT~\citep{S3_1_3-ViViT_2021}). This model takes Vision Transformer (ViT)~\citep{S1_13-ViT_2020} as its backbone and captures spatio-temporal features. ViViT has proven its value on Facial Expression Recognition (FER)~\citep{S1_2-ViViT-FER_2021} and violence detection~\citep{S1_3-ViViT-Vio_2022}.

The I-CONECT dataset has both inter- and intra-class variations, which makes it a challenging video dataset. First, it is an imbalanced dataset as the distributions of MCI and NC subjects are uneven. This makes the dataset inter-class imbalance (aka Hard-Easy sample problem). The intra-class imbalanced issue happens within each class as the I-CONECT dataset consists of videos in different lengths. Also, the quality of the videos in I-CONECT may vary from subject to subject. The occlusion and lightness problems in the videos decrease the quality of extracted spatio-temporal features. Furthermore, some subjects may not behave any symptoms of MCI, and act like NC in some video clips, which restricts ViViT from extracting enough MCI-related features. This is the so-called Positive-Negative sample problem.

To address these problems, we propose a Multi-branch Classifier (MC) module to augment the representation capability of ViViT. The proposed MC has four levels of Fully Connected (FC) layers, whereas the third one has 4 branches. MC provides more features but gives all features identical weight. Purely depending on MC contributes limitedly to the accuracy. We decide to assign different weights to features and classes while computing loss. Subsequently, we combined Focal loss~\citep{S2_4_3-Focal-Loss_2017} and AD-CORRE loss~\citep{S2_4_17-Adcorr_2022} into the loss function for Hard-Easy and Positive-Negative Samples (HP loss) to relief the negative effect of inter- and intra-class imbalanced issues. Specifically, the Focal Loss is responsible for the Hard-Easy sample problem, while AD-CORRE loss~\citep{S2_4_17-Adcorr_2022} handles the Positive-Negative sample problem. 

In summary, we present a new model called Multi-branch Classifier-ViViT (MC-ViViT), by integrating the aforementioned ViViT, MC, and HP Loss. We validated MC-ViViT on several themes of the I-CONECT dataset. Our experimental results show that MC-ViViT is highly capable to detect MCI from NC subjects. 

The overall contributions are summarized as follows:

\begin{itemize}
\item We propose MC-ViViT to detect MCI from the interview videos provided by the I-CONECT Study.
\item We design the MC module to enrich the extracted spatio-temporal features. Its multi-branch structure helps ViViT to capture the visual features from different perspectives.
\item We develop the HP loss by combining Focal loss and AD-CORRE loss. The HP loss addresses the inter- and intra-class imbalanced issues and helps the model pay attention to classes with less samples and subjects with short video lengths.
\end{itemize}

The remainder of this paper is organized as follows. In Section~\ref{sec:2}, we discuss the previous works on analyzing and recognizing facial expressions and patterns of MCI people in videos using machine learning approaches. We also review the methods for augmenting features and handling Inter- and Intra-class imbalanced problems. Then, in Section~\ref{sec:3}, we present our new proposed model, including Tubelet Embedding, Factorised Encoder, Multi-branch Classifier, and HP Loss. Section~\ref{sec:4} gives the details of our experiments including the dataset, data processing, evaluation metrics, implementation, results, discussion, and the ablation study. We finally conclude the paper with some suggestions for future research in Section~\ref{sec:5}.

\section{Related Work} \label{sec:2}

\subsection{Detection of Mild Cognitive Impairment (MCI) Using Machine Learning Methods} \label{sec:2.1}

Detection of MCI in older adults using innovative machine learning approaches has received great attention in the research community~\citep{S2_1_24-IADL_2020}. It is known that cognitive impairment and dementia adversely affect people's verbal and nonverbal behaviors, memory, cognitive functions, etc. Since the focus of this paper is on the detection of MCI using nonverbal visual patterns and behaviors expressed in facial videos of older adults using deep machine learning methods, we review the literature on using ML methods for the detection of MCI. 
Some researchers have utilized traditional ML approaches to detect MCI or more advanced stages of cognitive impairment such as Alzheimer's disease and dementia. ML methods such as Support Vector Machines (SVM), Decision Trees, PCA+SVM, K-means cluster, hierarchical clustering, and Density-based spatial clusters of applications with noise (DBSCAN)~\citep{S2_1_6-PCA_SVM_RBF_2018,S2_1_5-SVM_DTree_2020,S2_1_7-survey1_2020,S2_1_25-NACC_2023,S2_1_33-CNN-SVM-MRI_2023} are the most common algorithms used in this domain.

Some researchers have exploited deep learning models to detect MCI and dementia, such as CNN models with Inception modules~\citep{S2_1_8-Inception_2019}, linking a fully convolutional network (FCN) to a traditional multilayer perceptron (MLP)~\citep{S2_1_9-FCN_MLP_2020}, Inception-ResNet-V2~\citep{S2_1_10-Inception_ResNet_V2_2021}, and CNN-based models~\citep{S2_1_11-CNN_2020,S2_1_31-EffNets_2022,S2_1_32-DTI-CNN-RF_2021,S2_1_33-CNN-SVM-MRI_2023}.

Researchers have proposed that cognitive impairment and AD causes severe face recognition deficits and emotion detection deficits~\citep{S2_1_18-HFAD_2015,S2_1_13-ANCOVA_2018,S2_1_14-Happy_2018,S2_1_15-Compare_Mild_Moderate_2019,S2_1_12-ANOVA_2020,S2_1_16-Effect_of_Interference_2020,S2_1_17-Mask_COVID}. They collected the reaction of participants to specific emotions to estimate the degree of recognition deficits, which is a kind of MCI.

The aforementioned work did experiments on either brain Magnetic Resonance Imaging (MRI) scans, CT scans, and X-ray Scans \citep{S2_1_1-Early_Diagnosis_2018,S2_1_6-PCA_SVM_RBF_2018,S2_1_8-Inception_2019,S2_1_7-survey1_2020,S2_1_11-CNN_2020,S2_1_12-ANOVA_2020,S2_1_3-COVID19_2021,S2_1_4-ResUNet_2022,S2_1_10-Inception_ResNet_V2_2021,S2_1_31-EffNets_2022,S2_1_32-DTI-CNN-RF_2021,S2_1_33-CNN-SVM-MRI_2023} or demographic information, patient interviews, and structured datasets that are available on line~\citep{S2_1_18-HFAD_2015,S2_1_6-PCA_SVM_RBF_2018,S2_1_13-ANCOVA_2018,S2_1_14-Happy_2018,S2_1_15-Compare_Mild_Moderate_2019,S2_1_5-SVM_DTree_2020,S2_1_7-survey1_2020,S2_1_9-FCN_MLP_2020,S2_1_12-ANOVA_2020,S2_1_17-Mask_COVID,S2_1_2-ML_Compare_2022,S2_1_25-NACC_2023}. Collecting brain MRI and CT Scans are expensive. Other data modalities such as demographic information, patient interviews, and online structured datasets contain less complex, subtle, and subjective features. On the other hand, audiovisual collected during social interviews and face-to-face communication either in person or virtually contain rich verbal and nonverbal information.

Some researchers \citep{S2_1_19-EyeGaze_2020} video recorded the reaction of participants to different commands and introduced eye gaze and head pose in their study. \cite{S2_1_21-MoreSample_2022}~indicated the influence of eye-tracking on predicting cognitive impairment too. \cite{S2_1_22-Interact_2019}~filmed the video during the human-agent interaction and considered action units, eye gaze, and lip activity in the study. It asked three fixed queries: Q1) What’s the date today?, Q2) Tell me something interesting about yourself, Q3) How did you come here today? Then, \cite{S2_1_20-DynamicApproach_2019}~argued that extracting facial features in dynamic approaches benefits analyzing the evolution of facial expressions and getting spatio-temporal features. It also showed different facial feature extraction techniques, such as Geometric, Appearance, Holistic, and Local. Finally, \cite{S2_1_23-Xception_2021}~explored the capability of Xception and other CNN-based models to distinguish between people with cognitive impairment and those without dementia. Intuitively, our work explores the feasibility of an advanced deep learning model to predict MCI on the I-CONECT dataset.

\subsection{Facial Video Analysis Using Deep Neural Models} \label{sec:2.2}

Automated analysis of facial videos using deep machine learning and computer vision has been studied in the last decades. The applications vary from facial expression recognition (FER)~\citep{S2_2_2-PD_2020,S2_2_3-FER-GCN_2021,S2_2_22-FER-Video_2017}, surveillance~\citep{S2_2_11-Surveillance-1_2022,S2_2_12-Surveillance-YOLO-ViT_2021,S2_2_13-Surveillance-HRFR_2022}, pose estimation~\citep{S2_2_14-Pose-MFDNet_2022,S2_2_15-Pose-AffNet_2021}, head gesture recognition~\citep{S2_2_16-Head-gesture-AER_2022,S2_2_17-Head-gesture-HAFAR_2021,S2_2_18-Head-gesture-CNN_2022,S2_2_23-ATPN_2022}, medical applications~\citep{S2_2_19-Medical-PWP_2021,S2_2_20-Medical-STEADY-PD_2021,S2_2_21-Medical-Tel_2021}, among others. This section presents related works that used deep neural network-based algorithms for facial video analysis. 

When it comes to FER, some works used the traditional technique of converting videos into frames. For example, \cite{S2_2_1-MI_2021}~compressed videos into an apex frame by the encoding algorithm. \cite{S2_2_2-PD_2020}~simply cut videos into images, but it enriched features by using facial landmark points. Loading videos directly is another method. FER-GCN \citep{S2_2_3-FER-GCN_2021}, TimeSformer \citep{S2_2_8-TSM_2021}, and three-stream network \citep{S2_2_7-S3D_2022} simulated to analyze the video directly by taking multi frames as input. They stated that a sequence-based model helps extract temporal features. Based on multi frames, some researchers implemented a multi-frame optical flow method to get the difference between two frames~\citep{S2_2_6-t-s-CNN_2020,S2_2_5-TMSAU-Net_2021}. They claimed that multi-frame optical flow enriched temporal features. The aforementioned studies defended their idea with good experimental results. Furthermore, \cite{S2_2_2-PD_2020}~focused on diagnosing Parkinson's Disease (PD), which inspires us to believe that Deep Learning-based models are able to detect MCI.

Using more deep neural methods, Transformer-based models utilize modules to help capture spatio-temporal features in one stream, such as Convolutional Spatiotemporal Encoding layer from TimeConvNets~\citep{S2_2_4-TimeConvNets_2020} and Temporal Shift Module (TSM) from TimeSformer~\citep{S2_2_8-TSM_2021} and ViViT-B/16×2 FE~\citep{S3_1_3-ViViT_2021}~(A ViT-Base backbone with a tubelet size of $h\times w\times t = 16\times16\times2$, FE represents factorized encoder). Other works utilized spatial-only attention and temporal-only attention consecutively to extract spatio-temporal features in one branch~\citep{S2_2_8-TSM_2021,S2_2_9-VideoT_2021,S2_2_10-VSwinT_2022}.

On the other hand, CNN-based models usually take multi-stream structure to extract spatio-temporal features separately, such as deep temporal–spatial networks~\citep{S2_2_6-t-s-CNN_2020}, TMSAU-Net~\citep{S2_2_5-TMSAU-Net_2021}, and three-stream network~\citep{S2_2_7-S3D_2022}. These models take optical flow, frame difference, and motion vector methods to help create temporal information instead of learning from the raw images through the model. Models with 3D Convolutional operation can extract spatio-temporal information in one stream, but their large hyperparameter scale improves computational complexity.

Comprehensively, to avoid the extra operation and implement the model thoroughly and efficiently, this research selected ViViT FE as the backbone of the proposed model.

\subsection{Feature Enrichment using MLP Head}\label{sec:2.3}

Inception Net~\citep{S2_4_9-InceptionNet_2019} is a classic CNN model. It consists of repeated Inception Modules, which apply various scaled filters to extract features on the same layer and broaden the network. The advantage of the Inception Module is to enrich the features. Multi-models also help augment representation and improve accuracy. Moreover, some researchers have shown the effect of multi-models on MCI detection~\citep{S2_4_10-MM_AD_1_2019,S2_4_11-MM_AD_2_2019,S2_4_13-MM_AD_4_2020,S2_4_14-MM_AD_5_2021,S2_4_15-MM_AD_6_2022}. In addition, a good Fully Connected layer benefits the model performance, such as XnODR and XnIDR~\citep{S2_4_18-xnodr_2021}. The Inception Net and multi-models inspired the design of the Multi-branch Classifier (MC) (see Section \ref{sec:3.3}).

\subsection{Inter-class and Intra-class Imbalanced Problems} \label{sec:2.4}

Inter-class and intra-class imbalanced problems usually prevent ML models from predicting well. In \citep{S2_4_1-Shapelet_2019} the authors explained that the inter-class imbalanced problem occurs when the number of samples in some classes is much larger than those in other classes, which causes the misclassification of rare class examples. The intra-class imbalanced dataset means that a class consists of several sub-concepts or sub-clusters. Moreover, at least one of the concepts or clusters is represented by a significantly less number of samples than the others~\citep{S2_4_1-Shapelet_2019}. The number of features from each sub-concept are unequal, which weakens the performance of classifier~\citep{S2_4_2-survey_2021}.

Focal Loss is powerful to equalize inter-class imbalanced datasets~\citep{S2_4_3-Focal-Loss_2017}. Moreover, classes with more features are easier to detect than those with less features. This is also called the Hard-Easy sample problem. To address the intra-class imbalanced issue, researchers have proposed methods such as feature selection~\citep{S2_4_1-Shapelet_2019}, resampling~\citep{S2_4_1-Shapelet_2019,S2_4_4-DUBE_2021}, and designing loss function~\citep{S2_4_5-DDA-Loss_2020,S2_4_6-WCenter-Loss_2020,S2_4_7-DACL_2021,S2_4_17-Adcorr_2022}. For example, EPIMTS (early prediction on the imbalanced multivariate time series) fused feature selection and resampling, which calls MUDSG to resample based on the extracted core shapelets~\citep{S2_4_1-Shapelet_2019}. \cite{S2_4_4-DUBE_2021}~utilized Soft Hard Example Mining (SHEM) to re-balance the error density distribution. To achieve intra-class balancing, they assigned instance-wise sampling probabilities according to the prediction of the current ensemble model $F_{t-1}(\cdot)$~\citep{S2_4_4-DUBE_2021}. Other works focus on upgrading loss function. They proposed Discriminant Distribution-Agnostic loss (DDA loss)~\citep{S2_4_5-DDA-Loss_2020}, Weighted Center Loss (WCL)~\citep{S2_4_6-WCenter-Loss_2020}, Deep Attentive Center Loss (DACL)~\citep{S2_4_7-DACL_2021}, and Quadruplet loss~\citep{S2_4_8-Quadruplet-loss_2021}. These loss functions are tightly related to center loss. The goal of these methods is to achieve intra-class compactness and inter-class separation. The AD-CORRE loss presented in \citep{S2_4_17-Adcorr_2022} aims to handle inter- and intra-class imbalanced issues, but it focuses on addressing the problem by digging the correlation between embeddings in the mini-batch level. Then, it embeds the influence of whole training samples into batches incrementally instead of directly~\citep{S2_4_17-Adcorr_2022}. This has less computational cost.

In the I-CONECT dataset, the number of samples with MCI is apparently more than that with NC. Thus, subjects labeled in MCI are easy samples, while those labeled in NC are hard ones. To emphasize and increase the weight of NC, we apply Focal Loss to address the Hard-Easy sample problem. 

Within each class, the number of frames from each video is unequal. This is particularly important as interviewees with MCI may behave cognitively normal sometimes or show the symptoms of MCI in other parts of the video. The proportion of normal and symptom parts is imbalanced and skewed too. This can affect the accuracy of the final classification as we use majority voting to assign a video to MCI or NC based on the number of classified clips.

In general, the I-CONECT dataset has one inter-class imbalanced problem and two intra-class imbalanced problems. Based on the above discussion, we aim to achieve both inter- and intra-class separation, which is different from the target of the aforementioned loss functions such as DDA~\citep{S2_4_5-DDA-Loss_2020}, WCL~\citep{S2_4_6-WCenter-Loss_2020}, and DACL~\citep{S2_4_7-DACL_2021}. To avoid suboptimal performance, after comprehensive thought, we are inclined to use the Adaptive Correlation (AD-CORRE) loss~\citep{S2_4_17-Adcorr_2022} to solve the Positive-Negative sample problem.

\section{Multi-branch Classifier-ViViT (MC-ViViT)} \label{sec:3}

The proposed Multi-branch Classifier-ViViT (MC-ViViT) is an end-to-end model, which predicts whether a video segment in the I-CONECT dataset belongs to the MCI group or normal cognition (NC). The backbone of the network is ViViT FE~\citep{S3_1_3-ViViT_2021} as its Factorised Encoder structure can extract Spatio-Temporal features efficiently. To enrich the features and improve performance, MC-ViViT utilizes multi branches for classification. Specifically, MC-ViViT splits each video into several video cubes by Tubelet Embedding. Then, it embeds the video cubes into tokens and prepares the input tokens by concatenating class tokens and positional embedding. Subsequently, MC-ViViT applies the Transformer Encoder with Factorised Encoder to extract Spatio-Temporal features from the input tokens. Finally, the model computes the prediction score of each class using multi-branch classifiers. The rest of this section explains the model components in detail.
%\vspace{-0.4cm}

\subsection{Tubelet Embedding Review} \label{sec:3.1}

\begin{figure}[ht]
\begin{center}
\includegraphics[width=7.5cm,height=4.5cm]{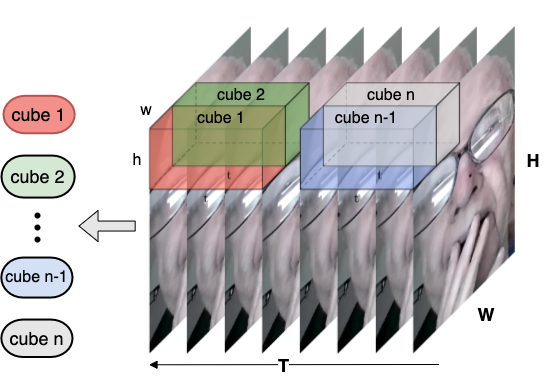}
\end{center}
\vspace{-0.5cm}
\caption{The structure of Tubelet Embedding.}
\label{fig:VVT-TE}
\end{figure}

Following the pattern of ViViT~\citep{S3_1_3-ViViT_2021}, the cubic patch is non-overlapping too. Suppose that the tensor shape of one video clip is $[T,H,W,3]$, where $T$ is the frame numbers, $H$ and $W$ are the height and the width of each frame, and $3$ represents the RGB channels. Then, the tensor size of each cubic patch is $[t,h,w,3]$. $t$, $h$, and $w$ are the size of the corresponding temporal, height, and width dimensions. $n_{t}=\lfloor\frac{T}{t}\rfloor$\footnotemark{}, $n_{h}=\lfloor\frac{H}{h}\rfloor$, and $n_{w}=\lfloor\frac{W}{w}\rfloor$ denote the token number of respective temporal, height, and width dimensions. The Tubelet Embedding changes the input unit from a 2D patch to a 3D cube, which contains temporal information (see Fig.~\ref{fig:VVT-TE}). Therefore, it is the foundation of offering Spatio-Temporal information in MC-ViViT.\footnotetext{$\left \lfloor \cdot/ \cdot \right \rfloor$ means to get the largest integer not greater than $\cdot/\cdot$.}

\subsection{Factorised Encoder (FE) Review} \label{sec:3.2}

\begin{figure}[ht]
\begin{center}
\includegraphics[width=8.cm,height=5.6cm]{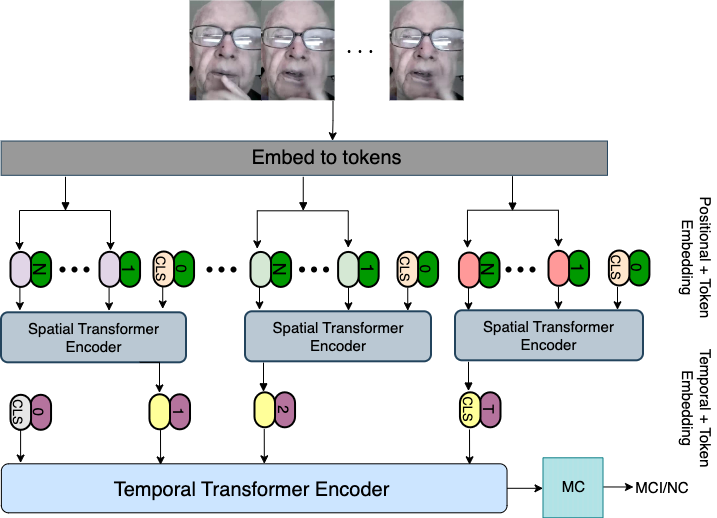}
\end{center}
\vspace{-0.6cm}
\caption{The structure of FE. CLS is class token. The green and purple capsules are positional embeddings. The rest capsules are tubelet embedding. MC is a Multi-branch classifier.}
\label{fig:VVT-FE}
\vspace{-0.4cm}
\end{figure}

Fig.~\ref{fig:m-struct} shows that in ViViT~\citep{S3_1_3-ViViT_2021}, the Transformer Encoder takes the combined positional and embedded cubic tokens as input, and it contains the Self-Attention module followed by the FeedForward module. Eq.~\textcolor{cyan}{(}\ref{eq:input-embed}\textcolor{cyan}{)} presents the details of the input tokens.

\vspace{-0.4cm}
\begin{equation} \label{eq:input-embed}
\mathbf{z}=[z_{cls},\mathbf{E}x_{1},\mathbf{E}x_{2},\dotsm,\mathbf{E}x_{n_{h}n_{w}n_{t}}]+\mathbf{p},
\end{equation}
where $z_{cls}$ is a learnable class token, $[x_{1},x_{2},\dotsm,x_{n_{t}n_{h}n_{w}}]$ is the tensor of cubic patches, $x_{i}\in \mathbb{R}^{h\times w\times t}$ $n_{h}n_{w}n_{t}$ is the number of cubic patches. $E$ is the linear projection to embed cubic patches. In addition, $\mathbf{p}\in \mathbb{R}^{(n_{h}n_{w}n_{t}+1)\times d}$ represents a learnable positional embedding. $d$ is the dimension of the embedded token.

Given the layer $l$, the Self-Attention module consists of Layer Normalization (LN) and Multi-Head Dot-Product Attention, while the FeedForward (FF) module includes LN and MLP. Both modules use skip-connection to enrich features and prevent gradient vanishing. Moreover, \cite{S3_1_3-ViViT_2021}~proposed four Multi-Head Self-Attention (MHSA) modules. This work selected FE because \cite{S3_1_3-ViViT_2021}~showed that ViViT FE performed the best of all four versions. Fig.~\ref{fig:VVT-FE} shows that FE has two parts, Spatial Transformer Encoder ($L_{s}$) and Temporal Transformer Encoder ($L_{t}$). Spatial Transformer Encoder extracts latent representations on the different tokens with the same temporal index. Therefore, these latent representations are spatial features. Then, these latent representations with different temporal indexes come to the Temporal Transformer Encoder, which studies the interactions between tokens from different time steps. Thus, the Temporal Transformer Encoder can mine temporal features. Assuming that the $L_{s}$ and $L_{t}$ repeats $n_{sp}$ and $n_{tp}$ times, respectively. FF processes the output of FE and returns the Sequence-Level Spatio-Temporal feature. The above explanation can be summarized as Eqs.~\ref{eq:fe.1}-\ref{eq:fe.4}.

\vspace{-0.4cm}
\begin{align}
\mathbf{y}^{l}_{s} &= L^{l}_{s_{n_{sp}}}(\dotsm L^{l}_{s_{1}}(LN(\mathbf{z}^{l}))\dotsm) \label{eq:fe.1}\\
\mathbf{y}^{l}_{t} &= L^{l}_{t_{n_{tp}}}(\dotsm L^{l}_{t_{1}}(\mathbf{y}^{l}_{s})\dotsm) \label{eq:fe.2}\\
\mathbf{y}^{l}_{FE} &= \mathbf{y}^{l}_{t} + \mathbf{z}^{l} \label{eq:fe.3}\\
\mathbf{y}^{l}_{FF} &= MLP(LN(\mathbf{y}^{l}_{FE})) + \mathbf{y}^{l}_{FE} \label{eq:fe.4}
\end{align}

\subsection{Multi-branch Classifier (MC)} \label{sec:3.3}

\begin{figure}[ht]
\begin{center}
\includegraphics[width=8.cm,height=5.3cm]{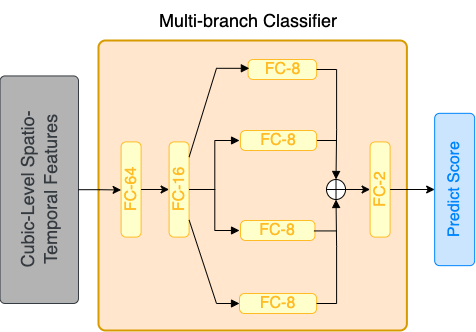}
\end{center}
\vspace{-0.4cm}
\caption{The structure of MC. $\oplus$ represents concatenation.}
\label{fig:VVT-AMLP}
\vspace{-0.5cm}
\end{figure}

Inspired by Inception Module~\citep{S2_4_9-InceptionNet_2019} and multi-models~\citep{S2_4_10-MM_AD_1_2019, S2_4_11-MM_AD_2_2019, S2_4_13-MM_AD_4_2020, S2_4_14-MM_AD_5_2021, S2_4_15-MM_AD_6_2022}, MC takes multi-branch structure to provide different views and enrich representation as well. Different from Inception Module and multi-models, MC only does linear projection at each branch instead of convolutional operation or the neural network. In detail, MC consists of four FC layers. The dimension change is $64\rightarrow 16\rightarrow [8,8,8,8]\rightarrow$ concatenate to $32\rightarrow num\_class$, where $num\_class$ represents the number of class. In our experiment, $num\_class$ is 2. When we convert the dimension from 16 to 32, Fig.~\ref{fig:VVT-AMLP} shows that we apply a multi-branch structure to convert the dimension to 8 and repeat 4 times. Then, we concatenate them as a 32-dimensional tensor. With this structure, MC can provide more features and view the object from different angles.

\subsection{Loss function for Hard-Easy and Positive-Negative Samples (HP Loss)} \label{sec:3.4}

HP Loss has two components, Focal Loss~\citep{S2_4_3-Focal-Loss_2017} and AD-CORRE(FD)~\citep{S2_4_17-Adcorr_2022}. AD-CORRE(FD) is the FD (Feature Discriminator) component of AD-CORRE Loss. 

\subsubsection{Focal Loss Review} \label{sec:3.4.1}

Focal Loss addresses the imbalance aspect of Hard-Easy samples by generating the weight based on sample numbers. It derives from $\alpha$-balanced Cross Entropy loss. Then, adding a modulating factor $(1-p_{mci})^{\gamma}$ defined in Eq.~\textcolor{cyan}{(}\ref{eq:pmci}\textcolor{cyan}{)} to the cross entropy loss, with tunable focusing parameter $\gamma>0$, makes the final Focal Loss. The above can be summarized as Eq.~\textcolor{cyan}{(}\ref{eq:focaloss}\textcolor{cyan}{)}.

\vspace{-0.4cm}
\begin{equation}
FL(p_{mci}) = -\alpha_{mci}(1-p_{mci})^{\gamma}log(p_{mci}) \label{eq:focaloss} 
\end{equation}
where $\alpha$ is a weighting factor. $\alpha\in[0,1]$ for class MCI and $1-\alpha$ for class NC. $p_{mci}$ represents the probability of the frame sequence attributing to MCI. 

\vspace{-0.4cm}
\begin{equation}\label{eq:pmci}
p_{mci} = \begin{cases} p & if\ y=MCI\\ 1-p & otherwise \end{cases} 
\end{equation}
where $p$ is the model's predict score, $y$ is the predicted label.

\subsubsection{AD-CORRE(FD) Review} \label{sec:3.4.2}

AD-CORRE(FD) focuses on the correlation between the samples within a mini-batch. This prevents the imbalanced dataset from affecting the prediction. Therefore, the model pays attention to the class with less samples too. AD-CORRE(FD) comprises four key components of AD-CORRE(FD), Variance Eraser (Beta Matrix $\beta_{n\times n}$), Attention Map Matrix ($\Omega_{n\times n}$), Harmony Matrix ($\Phi_{n\times n}$), and Correlation Matrix (CORM). In short, AD-CORRE(FD) is a weighted mean absolute error and is shown in Eq.~\textcolor{cyan}{(}\ref{eq:fd}\textcolor{cyan}{)}. 

\vspace{-0.4cm}
\begin{equation}\label{eq:fd}
\begin{split}
FD = & \frac{1}{kn^{2}}\displaystyle\sum\limits_{l=0}^k\displaystyle\sum\limits_{i=0}^n\displaystyle\sum\limits_{j=0}^n \beta[i,j]\Omega[i,j] \\
& |\Phi[i,j]-CORM_{l}[i,j]|, 
\end{split}
\end{equation}
where $k$ is the class number, $n$ is the size of mini-batch. The difference between $\Phi_{n\times n}$ and CORM$_{n\times n}$ is the core of the AD-CORRE(FD) loss.

In general, AD-CORRE(FD) focuses on the correlation between samples within the mini-batch. This prevents the imbalanced dataset from affecting the prediction. 

\textbf{AD-CORRE(FD) Analysis}

Intuitively, there are two ways to address the intra-class imbalance. Given a class, we assign trainable weight to each subject based on frame numbers. Or, we attribute different weights to positive and negative sequences within the same video. Either way, however, will stimulate more dispute and drive the problem complexity going exacerbation.

On this condition, AD-CORRE(FD)'s intention is to ignore the frames-imbalanced issue and the positive and negative sample problem and to only focus on the classification task at the mini-batch level. The intervention of AD-CORRE(FD) avoids the deeper conflict. In addition, AD-CORRE(FD) substantially decreases the computation complexity because it only analyzes the similarity between embeddings within the mini-batch. It accumulatively collects class distribution information from previous batches to improve its adaptive weight. Calculating within the mini-batch is also the reason that AD-CORRE(FD) benefits from detecting minority samples.

\subsubsection{Combine Two Loss Functions}

Finally, Eq.~\textcolor{cyan}{(}\ref{eq:HPLoss}\textcolor{cyan}{)} shows the HP Loss is the sum of Focal Loss and AD-CORRE (FD). We follow the pattern of AD-CORRE Loss and set $\lambda$ as $0.5$ as well.

\vspace{-0.4cm}
\begin{equation}
HP\ Loss = FL(p_{mci})+\lambda*AD-CORRE(FD) \label{eq:HPLoss}
\end{equation}

\section{Experiments} \label{sec:4}

In this section, we first introduce the I-CONECT dataset, evaluation metrics, and implementation details. Then, we explain our experiments, present the results, report the ablation study, and evaluate the results.

\subsection{Dataset} \label{sec:4.1}

The Internet-Based Conversational Engagement Clinical Trial (I-CONECT) is to explore how social conversation can help improve memory and may prevent dementia or Alzheimer’s disease in older adults. The study followed research participants aged 75 and older recruited from Portland, Oregon, or Detroit, Michigan in the USA. The study randomized 186 participants.  The participants had a 30-minute long chat per session with standardized interviewers (conversational staff) 4 times a week over a period of 6 months. The control group received only weekly 10-minute phone check-ins. All participants connected with the conversational staff using study-provided user-friendly devices. Conversations were semi-structured with some standard prompts and daily topics, but once the conversation started, it flowed naturally for fun and engaging conversations. Out of 186 randomized participants, 86 participants are diagnosed as NC, while 100 people are diagnosed as MCI. In each 30 minutes-session, participants discussed one of the 161 selected themes, such as Summer Time, Health Care, Military Service, Television Technology, etc. The conversational interactions were recorded as videoes. In the current study, we selected the following 4 themes: Crafts Hobbies, Day Time TV Shows, Movie Genres, and School Subjects. The interviewees talked about their crafts hobbies. They listed their favorite daytime TV shows, discussed preferred movie genres, and showed examples. Then, they also recalled their campus lives in School Subjects. Table~\ref{tab:theme_intro} shows data exploration.

We used sequence-based approaches rather than frame-based approaches to process the dataset and extracted facial features via a dynamic approach.

\begin{table} [ht]
\caption{The details of researched themes. Subject Number is the total number of videos in the corresponding theme. Male/Female and MCI/NC show the gender distribution and category distribution. Frame Number is the total image number of each theme after converting videos into frames.}
\vspace{-0.5cm}
\begin{center}
\resizebox{8.5cm}{!}{
\begin{tabular}{lcccc}
\hline
& \multirow{2}{1.5cm}{Crafts Hobbies} & \multirow{2}{1.6cm}{Day Time TV Shows} & \multirow{2}{1.5cm}{Movie Genres} & \multirow{2}{1.5cm}{School Subjects} \\
& & & & \\ 
\hline
Subject Number & 32 & 41 & 35 & 39 \\ 
Male/Female & 11/21 & 11/30 & 10/25 & 11/28 \\
MCI/NC & 20/12 & 20/21 & 21/14 & 22/17 \\
Frame Number & 691872 & 859872 & 770656 & 797968 \\
\hline
\end{tabular}}
\end{center}
\label{tab:theme_intro} \vspace{-0.7cm}
\end{table} 

\subsection{Data Processing}

\begin{figure}[ht]
\centering
\subfloat[]{{\includegraphics[width=4cm]{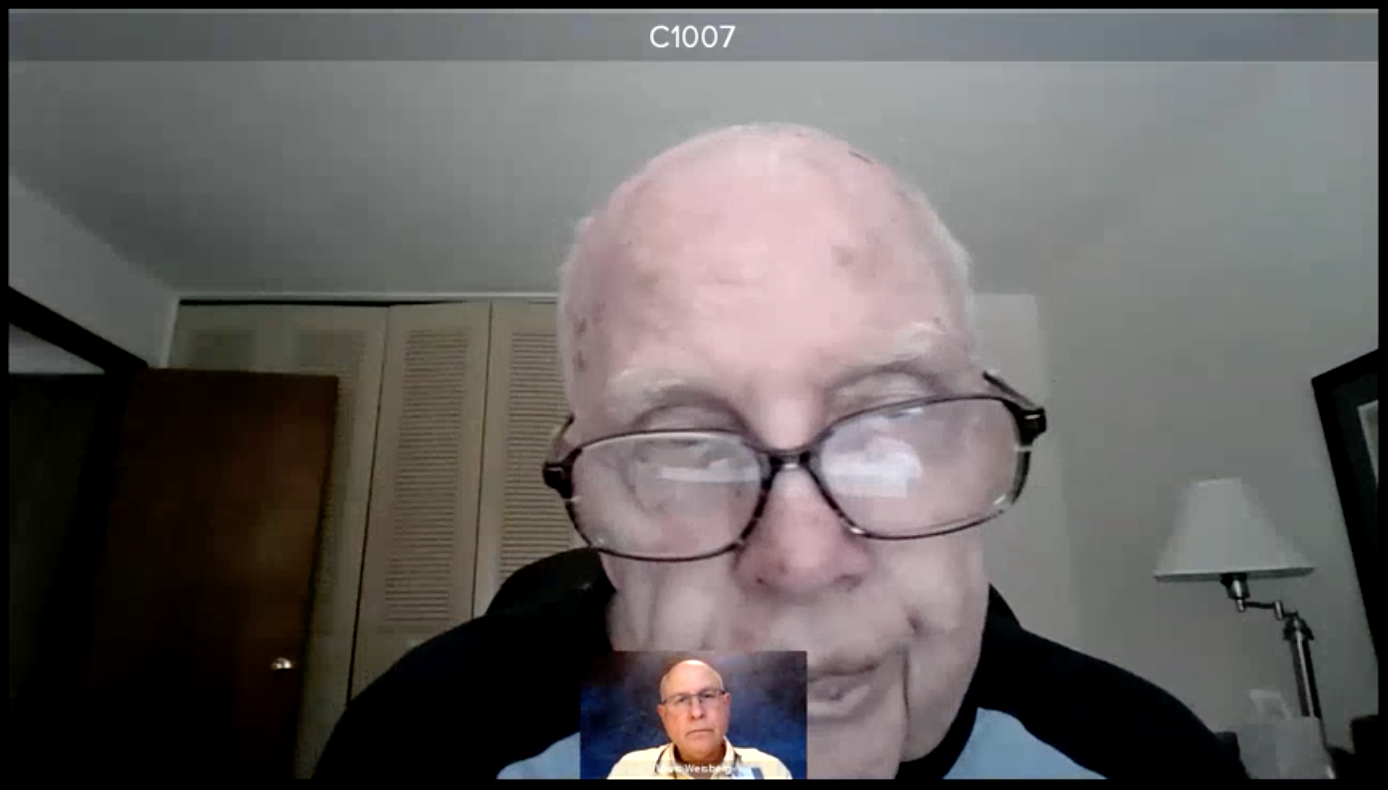} \label{fig:frame-l}}}
\subfloat[]{{\includegraphics[width=4cm]{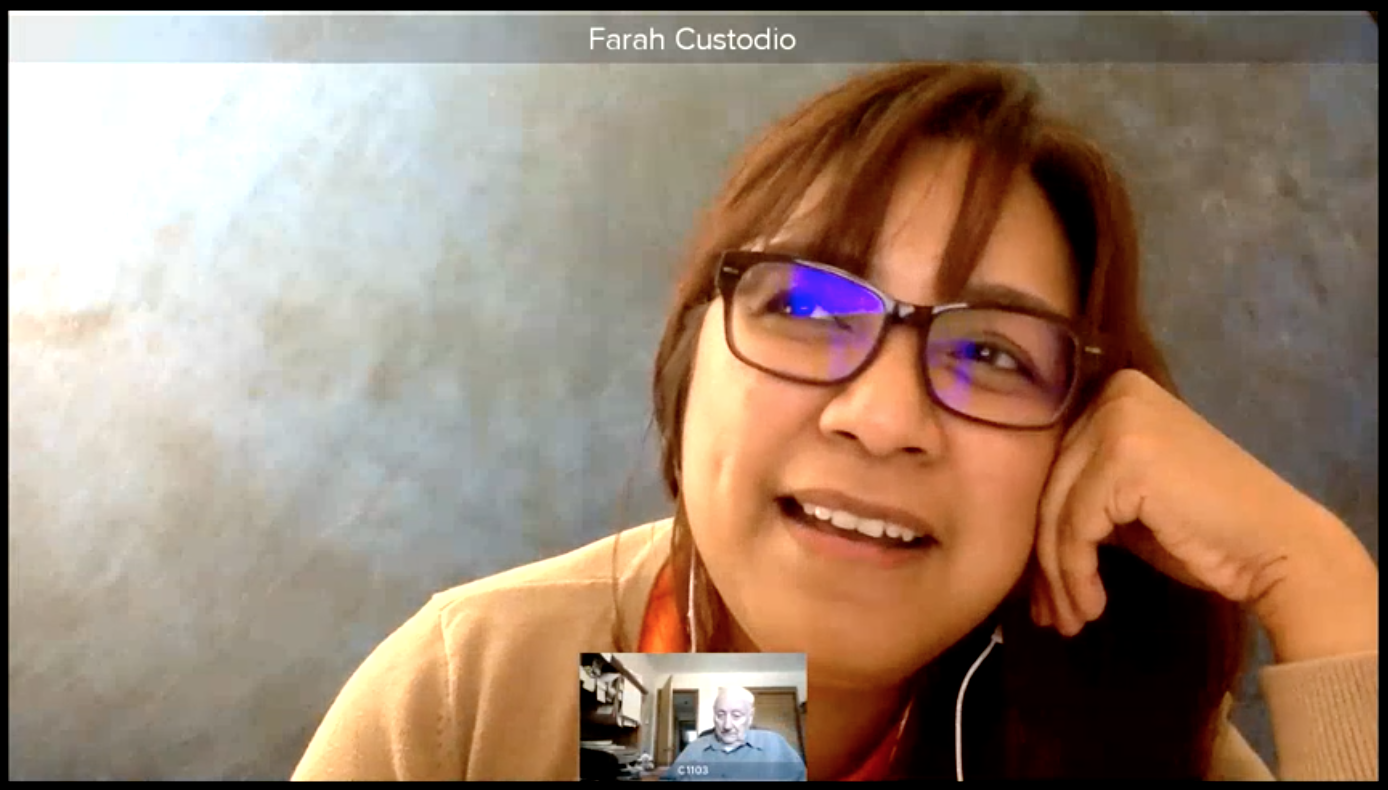}\label{fig:frame-s}}}
\vspace{-0.2cm}
\caption{Two sample frames from the video dataset. In (a), the window of the interviewee is bigger than that of the interviewer because the interviewer was speaking. Conversely, in (b), the interviewer was talking so that her window is bigger than the interviewee's.}
\label{fig:video-sample}
\vspace{-0.5cm}
\end{figure}

The facial features of interviewees are valuable to our research. Usually, the subject talks like NC at the start and the end, which prevents the model from predicting accurately. Furthermore, Fig. \ref{fig:video-sample} shows that every video has a complex background and interviewers' faces, which has a negative influence on predicting results. To address this problem, we decide to drop the first 3 minutes and the last 2.5 minutes of each video. Then, EasyOCR helps detect whether or not the upper half frame contains the subject ID in the given frame. For frames with the subject ID on the top, we implemented RetineFace~\citep{S4_2_1-retinaface_2020} to crop the participants' facial images and dropped all irrelevant and interfering information. In detail, we calculate the area of detected faces and the Intersection over Union (IoU) of them. If IoU $< 0.05$, we kept the bigger face like Fig.~\ref{fig:frame-l}. Otherwise, we continue to process the next frame and drop the frame like Fig.~\ref{fig:frame-s}. Finally, we browse the kept images and manually delete the wrongly saved ones. We used this method to process different themes. The four themes that are studied in this paper may have videos from 40 participants or less. For the sake of taking advantage of videos thoroughly and predicting better, we select K-fold Cross Validation and use Eq.~\textcolor{cyan}{(}\ref{eq:get_K}\textcolor{cyan}{)} to calculate $K$ for each theme.

\vspace{-0.4cm}
\begin{equation}\label{eq:get_K}
    K=\left \lfloor N_{Video}/L_{Fold} \right \rfloor 
\end{equation}
where $N_{Video}$ and $L_{Fold}$ are the video number of the selected theme and that of each fold. We set $L_{Fold}=3$ in our study. Moreover, videos in each fold belong to different participants. 

K-fold Cross Validation sets one fold as the test set and the resting folds as the training sets every time. It does not choose any fold as the validation set. Here, we use the test set as the validation one during the training process. This is to avoid overfitting the model~\citep{S4_2_2-MulBerry_Sort-2021}. In fact, attributing some of the subjects to a third subset inevitably shrinks the size of the training set. Thus, the model may face monotonous facial features and perform weakly in recognizing different samples~\citep{S4_2_2-MulBerry_Sort-2021}. Furthermore, the model may also take valuable representations as noise or outliers due to insufficient data samples, which adversely affects the prediction. Hence, we do not use a validation set in our experiments.

Simultaneously, inspired by~\cite{S2_2_4-TimeConvNets_2020, S2_2_3-FER-GCN_2021, S2_2_8-TSM_2021, S2_2_7-S3D_2022}, this study used multi frames as the input. Thus, we cut each video into a certain number of fixed-length segments as inputs. Let $L$\footnotemark{}\footnotetext{$L$ and $T$ from Section~\ref{sec:3.1} are identical.} be the number of consecutive frames in a divided segment, $N$ represents the number of total frames. $\left \lfloor N/L \right \rfloor$ is the number of segments for each video. 

\subsection{Evaluation Metrics}

Prediction accuracy, F1 score, AUC (Area Under the Receiver Operating Characteristic Curve), Sensitivity, and Specificity are the evaluation metrics in the experiment. McNamara and Martin stated that sensitivity and specificity are intrinsic measures of a case definition or diagnostic test, whereas predictive values vary with the prevalence of a condition within a population~\citep{S4_3_1-Sens_Spec-2023}. Eq.~\textcolor{cyan}{(}\ref{eq:sens}\textcolor{cyan}{)} presents the statistical definition of sensitivity, which is similar to recall. Eq.~\textcolor{cyan}{(}\ref{eq:spec}\textcolor{cyan}{)} is the equation of specificity.

\vspace{-0.4cm}
\begin{align}
\text{Sensitivity} &= \frac{\text{True Positive}}{\text{True Positive + False Negative}} \label{eq:sens} \\
\text{Specificity} &= \frac{\text{True Negative}}{\text{True Negative + False Positive}} \label{eq:spec}
\end{align}

\subsection{Implementation Details}

We performed data augmentation to each segment, which contains the random horizontal and vertical flip, random rotation, and center crop. All frames within one set do exactly same augmentation. The batch size is 80. Following the pattern of~\citep{S2_2_4-TimeConvNets_2020, S2_2_3-FER-GCN_2021,S2_2_8-TSM_2021,S2_2_7-S3D_2022}, $L$ is 16. The initialized learning rate is 1e-6. We used Adam optimizer and Cyclic scheduler with the mode of triangular2. The loss function is HP loss. The epoch number is 30. We coded the network by PyTorch 1.12.0+cu116 and ran experiments on the NVIDIA GTX 1080Ti GPU.

To completely evaluate the new proposed architecture, we designed three experiments (See Sections \ref{sec:4.5} - \ref{sec:4.7}).

\subsection{Experiment: K-fold Validation} \label{sec:4.5}

This experiment is to test the capability of MC-ViViT on predicting MCI in each theme. The inputs are all the corresponding videos. 

\begin{table}[ht]
\caption{The prediction accuracy of detecting MCI on 4 themes using K-fold evaluation.}
\vspace{-0.5cm}
\begin{center}
\resizebox{8.5cm}{!}{
\begin{tabular}{lcccc}
\hline
\multirow{2}{1.5cm}{Test Theme} & \multirow{2}{1.5cm}{Crafts Hobbies} & \multirow{2}{1.6cm}{Day Time TV Shows} & \multirow{2}{1.5cm}{Movie Genres} & \multirow{2}{1.5cm}{School Subjects} \\
& & & & \\ 
\hline
Fold Num & 11 & 14 & 11 & 13 \\ 
\multirow{2}{1.5cm}{Accuracy} & 29/32 & 36/41 & 30/35 & 35/39 \\
\cline{2-5}
& 90.63\% & 85.37\% & 85.71\% & 89.74\% \\
F1 & 93.03\% & 85.00\% & 88.37\% & 90.47\% \\
AUC & 0.6042 & 0.4952 & 0.6122 & 0.5294 \\
MCI/NC & 20/12 & 20/21 & 21/14 & 22/17 \\
Sensitivity/ & 100\%/ & 85\%/ & 90.48\%/ & 86.36\%/ \\
Specificity  & 75\% & 85.71\% & 78.57\% & 94.12\% \\
\hline
\end{tabular}}
\vspace{-0.6cm}
\end{center}
\label{tab:exp_one_theme}
\end{table} 

We report the results in Table~\ref{tab:exp_one_theme}. MC-ViViT achieves over 85\% accuracy on all four themes. For example, MC-ViViT predicts 29 out of 32 subjects correctly on the theme Crafts Hobbies, which reaches the highest accuracy, 90.63\%. The accuracy of School Subjects is 89.74\%, which is close to 90\%. Three of the F1 Scores are more than 88\%, and only the F1 score of Day Time TV Shows is 85\%. The accuracy and F1 Score solidly support that MC-ViViT has robust performance. The other evidence comes from AUC Score. The average AUC of all four themes is 0.56. Given that not all the clips have MCI features in the video, even for subjects with MCI, it is reasonable to predict parts of their video sequences as NC. Due to this reason, some subjects will be correctly predicted as MCI with low scores. Thus, the average AUC value is around 0.6. 

Table~\ref{tab:exp_one_theme} also shows that, on all four themes, MCI/NC has a related balanced prediction distribution. Their Sensitivity and Specificity are high as well. For instance, the sample of Day Time TV Shows comprises 20 MCI subjects and 21 NC ones. The Sensitivity over Specificity is 85\%/85.71\%.

\subsection{Experiment: cross themes evaluation} \label{sec:4.6}

This experiment is to test if MC-ViViT can integrate and learn complex features from a given set of training themes and detect MCI from different test themes correctly. We train the video on three themes and predict the rest. This time, we resize all the frames as $96\times 96$ to train faster. There are no overlapped subjects between the training sets and the test sets.

\begin{table}[ht]
\caption{The prediction accuracy of cross-theme fashion. Three themes are used for training a model and the left theme is used for test.}
\vspace{-0.5cm}
\begin{center}
\resizebox{8.5cm}{!}{
\begin{tabular}{lcccc}
\hline
\multirow{2}{1.5cm}{Test Theme}&\multirow{2}{1.5cm}{Crafts Hobbies} & \multirow{2}{1.6cm}{Day Time TV Shows} & \multirow{2}{1.5cm}{Movie Genres} & \multirow{2}{1.5cm}{School Subjects} \\
& & & & \\ 
\hline
Fold Num & 11 & 14 & 11 & 10 \\ 
\multirow{2}{1.5cm}{Accuracy} & 27/32 & 33/41 & 30/35 & 32/39 \\
\cline{2-5}
& 84.38\% & 80.49\% & 85.71\% & 82.05\% \\
F1  & 87.80\% & 82.61\% & 88.89\% & 89.36\% \\
AUC & 0.7     & 0.7452   & 0.6701  & 0.4773  \\
MCI/NC & 20/12 & 20/21 & 21/14 & 22/17 \\
Sensitivity/ & 90\%/ & 95\%/   & 95.24\%/ & 95.45\%/ \\
Specificity  & 75\%  & 66.67\% & 71.43\%  & 76.47\%  \\
\hline
\end{tabular}}
\end{center}
\label{tab:exp_cross_theme}
\vspace{-0.8cm}
\end{table} 

Table~\ref{tab:exp_cross_theme} shows all the results. MC-ViViT still provides stable performance. It achieves over 82\% accuracy and over 87\% F1 Score on the three themes. The highest accuracy is 85.71\% (predict 30 out of 35 samples correctly) on the Movie Genres, while the highest F1 Score is 89.36\% on the School Subject. The accuracy and F1 Score of this experiment are still robust to show that MC-ViViT can predict well. The average AUC Score of this experiment is 0.648, which is higher than 0.56 from Section~\ref{sec:4.5}. This value also accords with the analysis from Section~\ref{sec:4.5}. It supports that MC-ViViT can make objective decisions. In the meanwhile, MC-ViViT only reaches 80.49\% accuracy and 82.61\% F1 Score on the Day Time TV Shows. This is due to the smaller frame size.

\begin{figure*}[ht]
\centering
\subfloat[]{{\includegraphics[width=4cm]{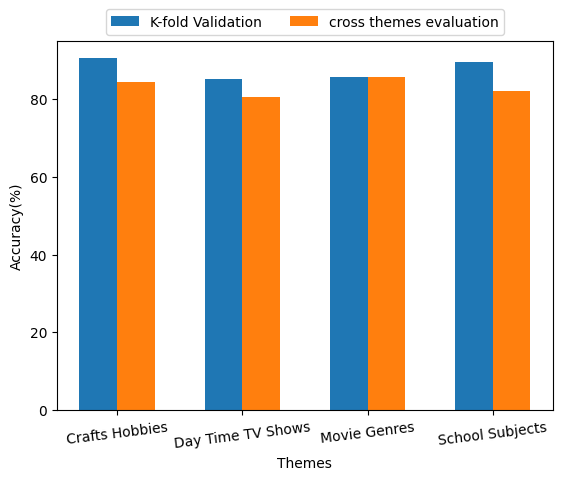} \label{fig:exp_KFold_Cross}}}
\subfloat[]{{\includegraphics[width=4cm]{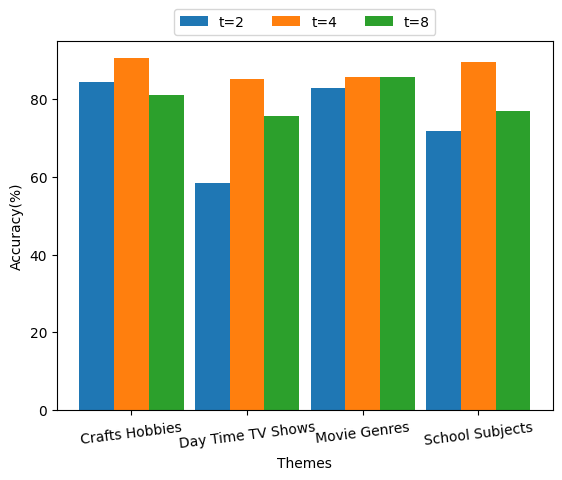}\label{fig:tune_t}}}
\subfloat[]{{\includegraphics[width=4cm]{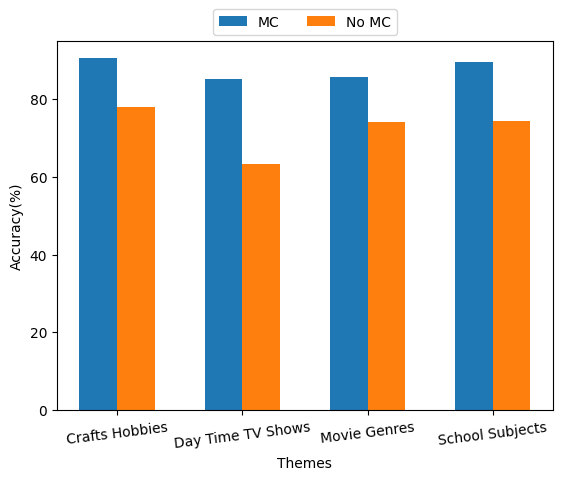} \label{fig:tune_MC}}}
\subfloat[]{{\includegraphics[width=4cm]{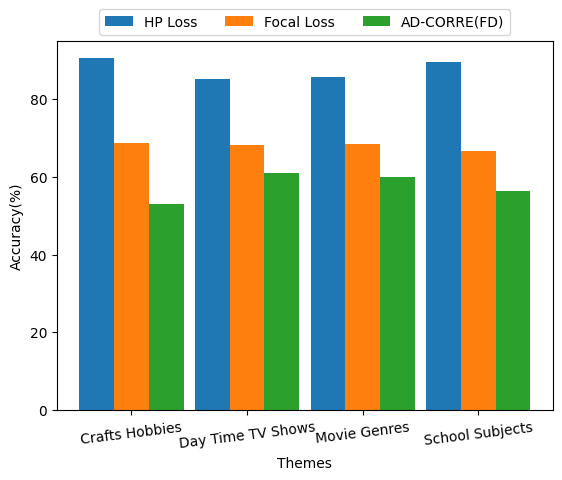}\label{fig:tune_Loss}}}
\vspace{-0.2cm}
\caption{(a): The prediction accuracy of MC-ViViT on four themes. The blue line represents the result of Section~\ref{sec:4.5}. Experiment: K-fold Validation, while the yellow one is that of Section~\ref{sec:4.6}. Experiment: cross themes evaluation. (b): The visualization of tuning temporal dimension $t$. $t=4$, MC-ViViT performs the best. (c): The visualization of tuning classifier MC. With MC, the model performs better. (d): The visualization of tuning loss function. With HP Loss, MC-ViViT performs better.}
\label{fig:plot_collect}
\vspace{-0.7cm}
\end{figure*}

Subsequently, Fig.~\ref{fig:exp_KFold_Cross} presents that, the smaller frame size affects the Sensitivity over Specificity of MCI/NC. Specifically, it weakens the capability of detecting negative samples, which is NC in our experiment. Thereby, the results of Sensitivity/Specificity are less balance than those from Section~\ref{sec:4.5}. For example, the Sensitivity/Specificity of movie genres increases from 90.48\%/78.57\% (Table~\ref{tab:exp_one_theme}) to 95.24\%/71.43\%. Smaller frame size decreasing the important features of NC makes the NC harder to detect. Therefore, Specificity in this experiment is lower than Section~\ref{sec:4.5}.

\subsection{Ablation Study} \label{sec:4.7}

This section discusses the effect of temporal dimension $t$, MC, and HP Loss on the performance of MC-ViViT (Sections \ref{sec:4.7.1}-\ref{sec:4.7.3}).

\subsubsection{Study the most proper temporal dimension $t$} \label{sec:4.7.1}

In MC-ViViT, the frame number, $T$, is 16, which can be divided by $[1,2,4,8,16]$. To keep Tubelet Embedding and reduce the burden of computation, we set the temporal dimension $t$ as $[2,4,8]$ alternatively to study the best value for this research.

\begin{table}[H]
\small
\caption{The prediction accuracy on 4 themes with different t size.}
\vspace{-0.5cm}
\begin{center}
\resizebox{8.5cm}{!}{
\begin{tabular}{lcccc}
\hline
\multirow{2}{.8cm}{Test Theme} & \multirow{2}{1.5cm}{Crafts Hobbies} & \multirow{2}{1.5cm}{Day Time TV Shows} & \multirow{2}{1.5cm}{Movie Genres} & \multirow{2}{1.5cm}{School Subjects} \\
& & & & \\ 
\hline
t = 2 & 84.38\% & 58.54\% & 82.86\% & 71.79\% \\ 
t = 4 & 90.63\% & 85.37\% & 85.71\% & 89.74\% \\
t = 8 & 81.25\% & 75.61\% & 85.71\% & 76.92\% \\
\hline
\end{tabular}}
\end{center}
\vspace{-0.5cm}
\label{tab:tune_t}
\end{table} 

Table \ref{tab:tune_t} and Fig.~\ref{fig:tune_t} show that setting $t=4$ gives the best performance under the current configuration on four themes. When $t=2$, the tubelet lacks adequate temporal features for making good predictions. When $t=8$, the tubelet, as a long sequence or long memory, includes many useful temporal representations for sure. But it also requires tubelet embedding to have a larger fixed dimension length. However, increasing this fixed length will boost the model's total parameters, which makes the MC-ViViT computationally expensive.

In addition, other Transformer-based models also set $t=4$ while analyzing facial videos such as MTV~\citep{S4_7_1_6-MTV-2022} and MSVT~\citep{S4_7_1_7-MSVT-2023}. Using $t=4$ helps Transformer-based models capture local-consecutive inconsistency efficiently and reach sufficient performance~\citep{S4_7_1_7-MSVT-2023}. It is also to get the best trade-off between performance and efficiency~\citep{S4_7_1_6-MTV-2022}. Thereby, we choose $t=4$.

\subsubsection{Study the effect of MC} \label{sec:4.7.2}

This study is to evaluate the effect of MC structure. In the experiment, we dropped the multi-branch structure and changed the dimension from 16 to $num\_class$ directly. The dimension changes will be $64\rightarrow 16\rightarrow num\_class$. We collected the related experimental results in Table~\ref{tab:tune_AM}. It shows that, for each theme, the accuracy of ViViT with MC is at least 10\% higher than that without MC. Fig.~\ref{fig:tune_MC} bolsters this view. Therefore, the MC module provides more features and benefits the ViViT in predicting better. 

\begin{table}[H]
\caption{The prediction accuracy on 4 themes with or without MC. No multi-branch means that we drop the multi-branch during the experiment.}
\vspace{-0.5cm}
\begin{center}
\resizebox{8.5cm}{!}{
\begin{tabular}{lcccc}
\hline
\multirow{2}{.8cm}{Test Theme} & \multirow{2}{1.5cm}{Crafts Hobbies} & \multirow{2}{1.6cm}{Day Time TV Shows} & \multirow{2}{1.5cm}{Movie Genres} & \multirow{2}{1.5cm}{School Subjects} \\
& & & & \\ 
\hline
MC & 90.63\% & 85.37\% & 85.71\% & 89.74\% \\ 
No MC & 78.13\% & 63.41\% & 74.29\% & 74.36\% \\
\hline
\end{tabular}}
\end{center}
\vspace{-0.5cm}
\label{tab:tune_AM}
\end{table} 

\subsubsection{Study the effect of HP Loss} \label{sec:4.7.3}
This study is to test the influence of each component of HP Loss on the prediction. We first did experiments with Focal Loss only. Then, we did experiments with AD-CORRE(FD) Loss only. Table~\ref{tab:tune_LOSS} contains all the results and shows that MC-ViViT using Focal loss performs better than that using AD-CORRE(FD). In the meanwhile, the accuracy of MC-ViViT using HP loss surpasses at least 14\% than that using either Focal loss or AD-CORRE(FD). Fig.~\ref{fig:tune_Loss} clearly displays this huge disparity. Thereby, Table~\ref{tab:tune_LOSS} and Fig.~\ref{fig:tune_Loss} uphold that HP Loss helps the MC-ViViT perform better than each individual component does.

\begin{table}[ht]
\caption{The prediction accuracy on 4 themes with different loss functions.}
\vspace{-0.5cm}
\begin{center}
\resizebox{8.5cm}{!}{
\begin{tabular}{lcccc}
\hline
\multirow{2}{2cm}{Test Theme} & \multirow{2}{1.5cm}{Crafts Hobbies} & \multirow{2}{1.6cm}{Day Time TV Shows} & \multirow{2}{1.5cm}{Movie Genres} & \multirow{2}{1.5cm}{School Subjects} \\
& & & & \\ 
\hline
HP Loss & 90.63\% & 85.37\% & 85.71\% & 89.74\% \\ 
Focal Loss & 68.75\% & 68.29\% & 68.57\% & 66.67\% \\
AD-CORRE(FD) & 53.13\% & 60.98\% & 60.00\% & 56.41\% \\
\hline
\end{tabular}}
\end{center}
\vspace{-0.5cm}
\label{tab:tune_LOSS}
\end{table} 

\section{Discussion} \label{sec:5}

The experiments presented in Sections~\ref{sec:4.5}~and \ref{sec:4.6}~and the comparison in Table~\ref{tab:otherWork}~indicate that MC-ViViT can detect MCI with promising accuracy. For a selected theme, MC-ViViT can predict subjects not included in the model training very well. It can also infer the representations learned from many themes to correctly predict the subjects from unseen themes. The MC-ViViT's robust performance convinces us that analyzing semi-structured interview videos is useful to detect MCI. 

\begin{table}[ht]
\caption{\cite{S5_1-Conv-2020} and \cite{S5_2-Lang-MRI-2022} mostly focused on language and medical image data in detecting MCI using the I-CONECT dataset, while we are one of the first groups to use non-medical visual data. These two papers have tried different subsets of I-CONECT and different data modalities. Nevertheless, our model (MC-ViViT) outperforms their work.}
\vspace{-0.5cm}
\begin{center}
\resizebox{8.3cm}{!}{
\begin{tabular}{lccc}
\hline
 & Data Modality & Accuracy & F1 \\
\hline
\multirow{1}{2.6cm}{\cite{S5_1-Conv-2020}} & Text & 79.15\% & - \\ 
\multirow{1}{2.6cm}{\cite{S5_2-Lang-MRI-2022}} & Text, MRI & 87.00\% & 89.00\% \\
MC-ViViT (Ours) & Video & \textbf{90.63\%} & \textbf{93.03\%} \\
\hline
\end{tabular}}
\end{center}
\vspace{-0.8cm}
\label{tab:otherWork}
\end{table} 

\begin{figure}[ht]
\vspace{-0.3cm}
\begin{center}
\includegraphics[width=6cm,height=6cm]{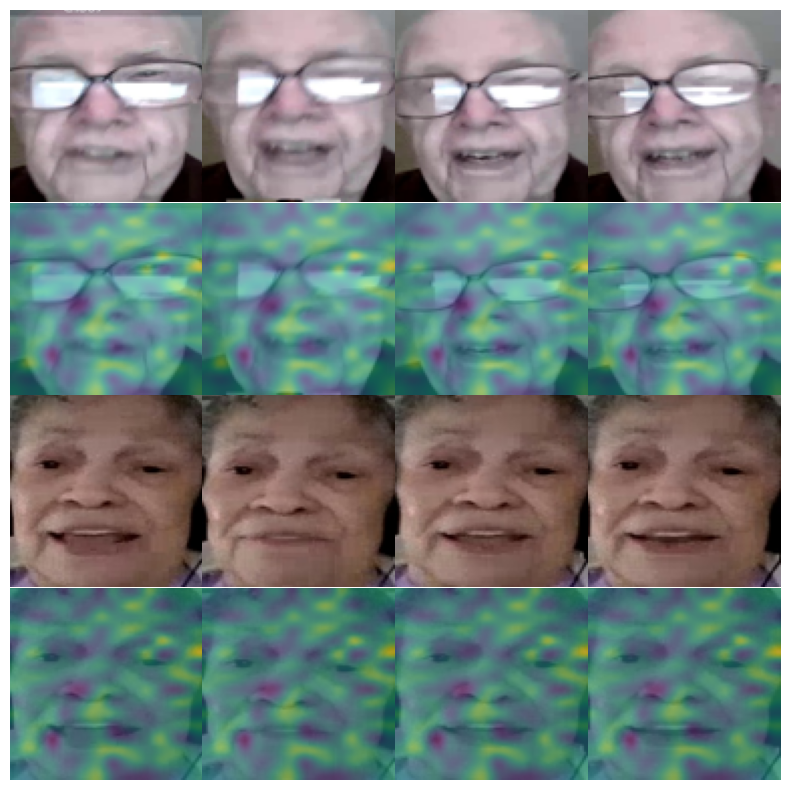}
\end{center}
\vspace{-0.3cm}
\caption{The extracted spatio-temporal feature maps on 16 consecutive frames from two subjects. MC-ViViT focuses more on the bright areas and pays less attention on the dark ones. Given that there are $n_{t}$ tubelets, where $n_{t}=\lfloor\frac{T}{t}\rfloor=\lfloor\frac{16}{4}\rfloor=4$, there are four feature maps. To simplify the visualization, we pick the first frame of each tubelet to represent it.}
\label{fig:attnmap}
\vspace{-0.5cm}
\end{figure}

In the meanwhile, ViViT, as the backbone, mainly provides good representations to the final prediction. These good representations include spatial features and temporal ones. Different from manually described and learned patterns such as head pose, eye gaze, eye-tracking, and lip activities, Fig.~\ref{fig:attnmap} shows that MC-ViViT pays attention to the different areas of the face. For example, it focuses on regions of the forehead, eyelids, nose, cheek, and jaw. This correlates with the statement that ViViT is capable to monitor the motion of all facial muscles and discover many slight changes and features. Furthermore, Sections~\ref{sec:4.5} and \ref{sec:4.6} indicate that these features are very important to decide if the subject has MCI or not. In short, ViViT takes video clips as the input and extracts spatio-temporal features in one stream, which enables the model to make a comprehensive decision. Subsequently, the ablation study from Section~\ref{sec:4.7} suggests that both MC and HP Loss improve the accuracy. MC broadens the network through the multi-branch structure, which provides different perspectives to analyze the subject's features. MC strengthens the features' richness, which benefits in making correct decisions. HP Loss, then, concentrates on prediction equality. The I-CONECT dataset is a very imbalanced dataset. Table~\ref{tab:theme_intro} indicates that every theme has more MCI subjects than NC ones. The focal Loss part of HP loss assigns more weight to the class with less subjects and aims to solve the inter-class imbalanced issues. Within each class, the different video length causes the inequivalence of extracted frames from each subject. The AD-CORRE(FD) part of HP Loss changes the background of the problem from the intra-class level to the mini-batch level. Within the mini-batch, AD-CORRE(FD) tackles the classification task by measuring the similarity between embedding features, which wisely resolves the imbalance crisis and reduces the computational complexity. Thereby, HP loss is essential to the MC-ViViT. The experimental results from Tables~\ref{tab:tune_AM} and~\ref{tab:tune_LOSS} uphold the above discussion. In addition, Table~\ref{tab:exp_one_theme} indicates that MC-ViViT can learn features well under one particular theme, while Table~\ref{tab:exp_cross_theme} convinces us that features from other themes can help detect MCI from NC on the left one. Subjects perform similar features on different themes.

In general, there are two key findings. The first one is that Transformer-based models are capable to detect MCI in the early stage only by analyzing facial features; the second one is that semi-structured interview videos can provide plenty of features to help the model make quality predictions while merely requiring the tablet computer or mobile phone to collect. This costs much less than collecting conventional medical images like MRI. These two findings reveal that MC-ViViT has the potential to detect other diseases, such as dementia and Parkinson's Disease (PD), in the early stage. Moreover, the corresponding data collection becomes convenient and easy to access. Even a cell phone can help film facial videos.

On the other side, due to the lack of a consistent video recording environment, the videos' qualities vary a lot. From what we observed in this research, the video quality significantly affects the prediction. We notice that MC-ViViT usually predicts high-clarity videos rightly with high prediction scores. Conversely, MC-ViViT is easy to give unclear videos with low prediction scores. Or even MC-ViViT classifies those videos with the wrong labels. This explains the reason that Sensitivity versus Specificity is far from equivalence. Future work will focus on establishing a mechanism to evaluate the video quality score. The future model will implement this score to design a new loss function that helps tune the training process.

\section{Conclusion} \label{sec:6}

This paper presented MC-ViViT to detect MCI from NC via facial videos of the semi-structured interviews (I-CONECT dataset). The experiments show that MC-ViViT can predict well with the help of MC and HP Loss. Simultaneously, it shows that the Transformer-based model can tell whether or not people have MCI merely by analyzing facial features from the semi-structured interview videos. Compared to the other datasets mentioned in Section~\ref{sec:1}, the I-CONECT dataset is easy to access, low cost, time-saving, flexible, and less restricted. It is very valuable to spread and propagate.

\section*{Acknowledgement}

This research was partially funded by a grant to the University of Denver by Colorado Office of Economic Development and International Trade and two grants from National Institutes of Health (NIH) (R01AG051628 and R01AG056102).

We would like to thank Miss. Jisu Lee for proofreading this paper.

\section*{Author Contributions}

\textbf{Jian Sun:} Conceptualization, Methodology, Software, Validation, Formal analysis, Investigation, Data Curation, Writing - Original Draft, Writing - Review \& Editing, Visualization.

\textbf{Hiroko H. Dodge:} Investigation, Resources, Data Curation, Writing---Review and Editing. 

\textbf{Mohammad H. Mahoor:} Resources, Writing---Review and Editing, Supervision, Project Administration.

%% Loading bibliography style file
%\bibliographystyle{model1-num-names}
%\bibliographystyle{model5-names}\biboptions{authoryear}

\bibliographystyle{model5-names}
\bibliography{cas-refs}

\end{document}